\documentclass[letterpaper, 10 pt, conference]{ieeeconf}  % Comment this line out
\IEEEoverridecommandlockouts                              % This command is only
\usepackage{cite}
\usepackage{graphics} % for pdf, bitmapped graphics files
\usepackage{epsfig} % for postscript graphics files
\usepackage{epstopdf}
\graphicspath {{./image/}}
\usepackage{mathptmx} % assumes new font selection scheme installed
\usepackage{times} % assumes new font selection scheme installed
\usepackage{amsmath} % assumes amsmath package installed
\usepackage{amssymb}  % assumes amsmath package installed
\usepackage{gensymb}
\usepackage{multirow}
\usepackage{picinpar}
\usepackage{multicol}
\usepackage{stfloats}
\usepackage{caption}
\usepackage{graphicx}
\usepackage{subfigure}
\usepackage{wasysym}
\usepackage{textcomp}
\usepackage[colorlinks,linkcolor=red,anchorcolor=blue,citecolor=green]{hyperref}

\title{\LARGE \bf
UnDeepVO: Monocular Visual Odometry through Unsupervised Deep Learning
}

\author{Ruihao Li$^{1}$, Sen Wang$^{2}$, Zhiqiang Long$^{3}$ and Dongbing Gu$^{1}$% <-this % stops a space
\thanks{$^{1}$Ruihao Li, Dongbing Gu are with School of Computer Science and Electronic Engineering, University of Essex, Colchester, CO4 3SQ, UK.
       {\tt\small \{rlig, dgu\}@essex.ac.uk}}%
\thanks{$^{2}$Sen Wang is with Edinburgh Centre for Robotics, Heriot-Watt University, Edinburgh, EH14 4AS, UK.  {\tt\small s.wang@hw.ac.uk}}%
\thanks{$^{3}$Zhiqiang Long is with College of Mechatronics and Automation, National University of Defense Technology, Changsha, China. }%
}

\begin{document}

\maketitle
\thispagestyle{empty}
% * <dongbinggu@gmail.com> 2017-09-15T15:36:54.560Z:
%
% ^.
% * <dongbinggu@gmail.com> 2017-09-15T15:36:51.767Z:
%
% ^.
\pagestyle{empty}

%%%%%%%%%%%%%%%%%%%%%%%%%%%%%%%%%%%%%%%%%%%%%%%%%%%%%%%%%%%%%%%%%%%%%%%%%%%%%%%%
\begin{abstract}

We propose a novel monocular visual odometry (VO) system called UnDeepVO in this paper. UnDeepVO is able to estimate the 6-DoF pose of a monocular camera and the depth of its view by using deep neural networks. There are two salient features of the proposed UnDeepVO: one is the unsupervised deep learning scheme, and the other is the absolute scale recovery. Specifically, we train UnDeepVO by using stereo image pairs to recover the scale but test it by using consecutive monocular images. Thus, UnDeepVO is a monocular system. The loss function defined for training the networks is based on spatial and temporal dense information. A system overview is shown in Fig. \ref{fig:front_fig}. The experiments on KITTI dataset show our UnDeepVO achieves good performance in terms of pose accuracy.

%It is different from geometry based VO methods as it is purely data-driven without using hand-designed feature extraction, etc. It can be trained in an unsupervised end-to-end manner leveraging geometry constraints encapsulated among images. Specifically, the absolute scales of the depth images and 6-DoF poses are recovered by minimizing the spatial image losses between stereo image pairs, while the 6-DoF poses are estimated through minimizing the photometric loss and 3D geometric registration loss between consecutive monocular images. Since only monocular imagery is needed to perform the VO after training, UnDeepVO is a monocular VO system. Extensive experiments on KITTI dataset show UnDeepVO outperforms other monocular VO methods in terms of accuracy.

\end{abstract}

%%%%%%%%%%%%%%%%%%%%%%%%%%%%%%%%%%%%%%%%%%%%%%%%%%%%%%%%%%%%%%%%%%%%%%%%%%%%%%%%
\section{Introduction}

Visual odometry (VO) enables a robot to localize itself in various environments by only using low-cost cameras. In the past few decades, model-based VO or geometric VO has been widely studied and its two paradigms, feature-based method\cite{davison2007monoslam,klein2007parallel,mur2015orb} and direct method\cite{newcombe2011dtam,engel2014lsd,engel2017direct}, have both achieved great success. However, model-based methods tend to be sensitive to camera parameters and fragile in challenging settings, e.g., featureless places, motion blurs and lighting changes.

In recent years, data-driven VO or deep learning based VO has drawn significant attention due to its potentials in learning capability and the robustness to camera parameters and challenging environments. Starting from the relocalization problem with the use of supervised learning, Kendall et al.\cite{kendall2015posenet} first proposed to use a Convolutional Neural Network (CNN) for 6-DoF pose regression with raw RGB images as its inputs. Li et al.\cite{li2017indoor} then extended this into a new architecture for raw RGB-D images with the advantage of facing the challenging indoor environments. Video clips were employed in \cite{clark2017vidloc} to capture the temporal dynamics for relocalization. Given pre-processed optical flow, a CNN based frame-to-frame VO system was reported in \cite{costante2016exploring}. Wang et al. \cite{wang2017deepvo} then presented a Recurrent Convolutional Neural Network (RCNN) based VO method resulting in a competitive performance against model-based VO methods. Ummenhofer \cite{ummenhofer2016demon} proposed ``DeMoN'' which can simultaneously estimate the camera's ego-motion, image depth, surface normal and optical flow. Visual inertial odometry with deep learning was also developed in \cite{clark2017vinet} and \cite{pillai2017towards}. 

	%TODO:REPLACE 2 DEPTH MAPs to RED IMAGES. 
    %TODO:CHANGE FIGURE LEGEND
	\begin{figure}
		\centering
		\includegraphics[width=1.0\columnwidth]{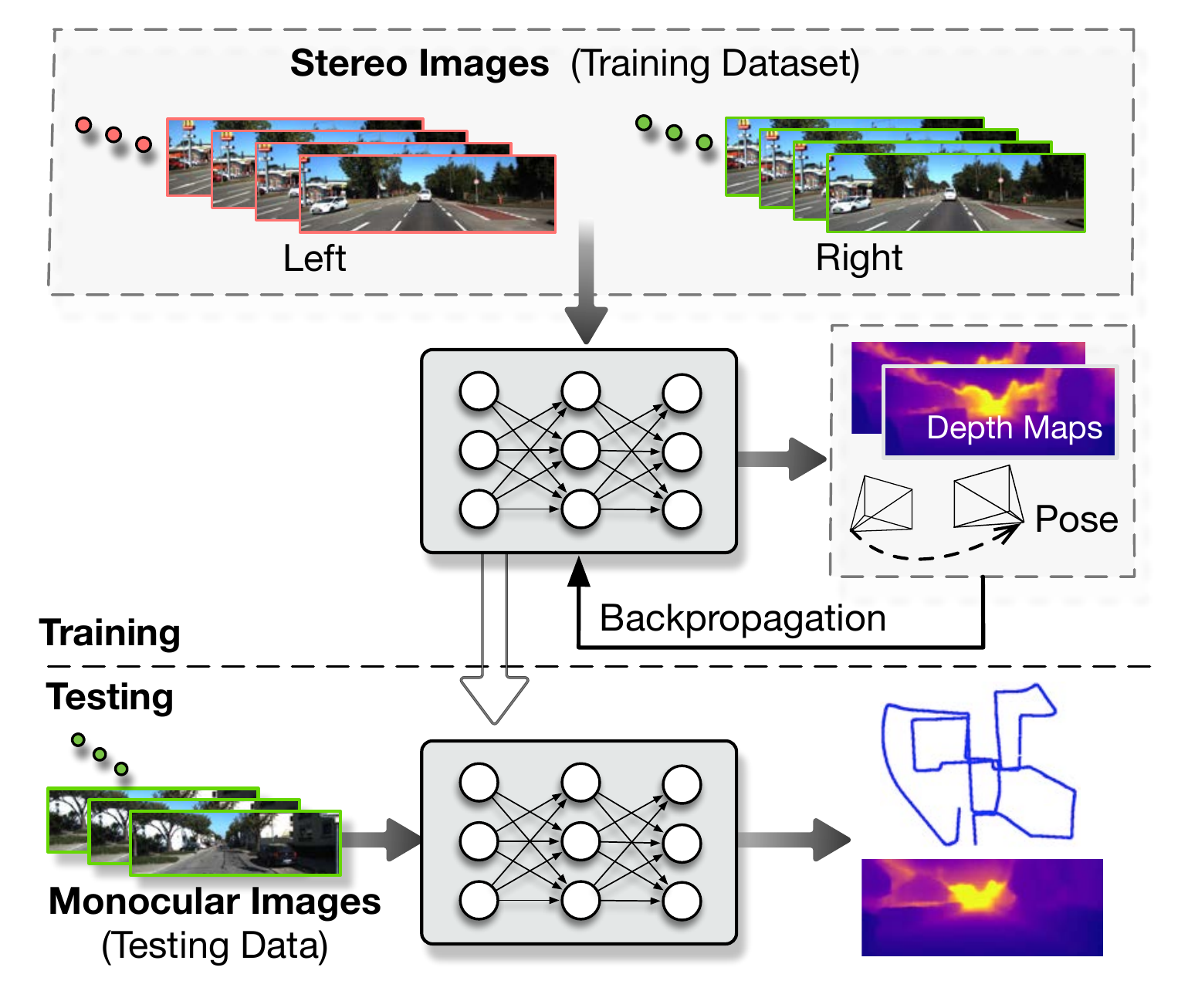}
		\caption{System overview of the proposed UnDeepVO. After training with unlabeled stereo images, UnDeepVO can simultaneously perform visual odometry and depth estimation with monocular images. The estimated 6-DoF poses and depth maps are both scaled without the need for scale post-processing.}
		\label{fig:front_fig}
	\end{figure}
    
However, all the above mentioned methods require the ground truth of camera poses or depth images for conducting the supervised training. Currently obtaining ground truth datasets in practice is typically difficult and expensive, and the amount of existing labeled datasets for supervised training is still limited. These limitations suggest us to look for various unsupervised learning VO schemes, and consequently we can train them with easily collected unlabeled datasets and apply them to localization scenarios. 

VO related unsupervised deep learning research mainly focuses on depth estimation, inspired by the image wrap technique ``spatial transformer'' \cite{jaderberg2015spatial}. Built upon it, Garg et al.\cite{garg2016unsupervised} proposed a novel unsupervised depth estimation method by using the left-right photometric constraint of stereo image pairs. This method was further improved in \cite{monodepth17} by wrapping the left and right images across each other. In this way, the accuracy of depth prediction was improved by penalizing both left and right photometric losses. Instead of using stereo image pairs, Zhou et al.\cite{zhou2017unsupervised} proposed to use consecutive monocular images to train and estimate both ego-motion and depth, but the system cannot recover the scale from learning monocular images. Nevertheless, these unsupervised learning schemes have brought deep learning technologies and VO methods closer and showed great potential in many applications.

In this paper, we propose UnDeepVO, a novel monocular VO system based on unsupervised deep learning scheme (see Fig. \ref{fig:front_fig}). Our main contributions are as follows:
\begin{itemize}
  \item We demonstrate a monocular VO system with recovered absolute scale, and we achieve this in an unsupervised manner by harnessing both spatial and temporal geometric constraints.
  \item Not only estimated pose but also estimated dense depth map are generated with absolute scales thanks to the use of stereo image pairs during training. 
  \item We evaluate our VO system using KITTI dataset, and the results show UnDeepVO achieves good performance in pose estimation for monocular cameras.
\end{itemize}

Since UnDeepVO only requires stereo imagery for training without the need of labeled datasets, it is possible to train it with an extremely large number of unlabeled datasets to continuously improve its performance.  
% Different from model-based VO methods, DL methods are data-driven based. ImageNet-scale SLAM dataset is almost impossible to obtain when requiring the ground-truth. 

The rest of this paper is organized as follows. Section \uppercase\expandafter{\romannumeral2} introduces the architecture of our proposed system. Section \uppercase\expandafter{\romannumeral3} describes different types of losses used to facilitate the unsupervised training of our system. Section \uppercase\expandafter{\romannumeral4} presents experimental results. Finally, conclusion is drawn in Section \uppercase\expandafter{\romannumeral5}.
%%%%%%%%%%%%%%%%%%%%%%%%%%%%%%%%%%%%%%%%%%%%%%%%%%%%%%%%%%%%%%%%%%%%%%%%%%%%%%%%

\section{System Overview}
\begin{figure}
  \centering
  \includegraphics[width=\columnwidth]{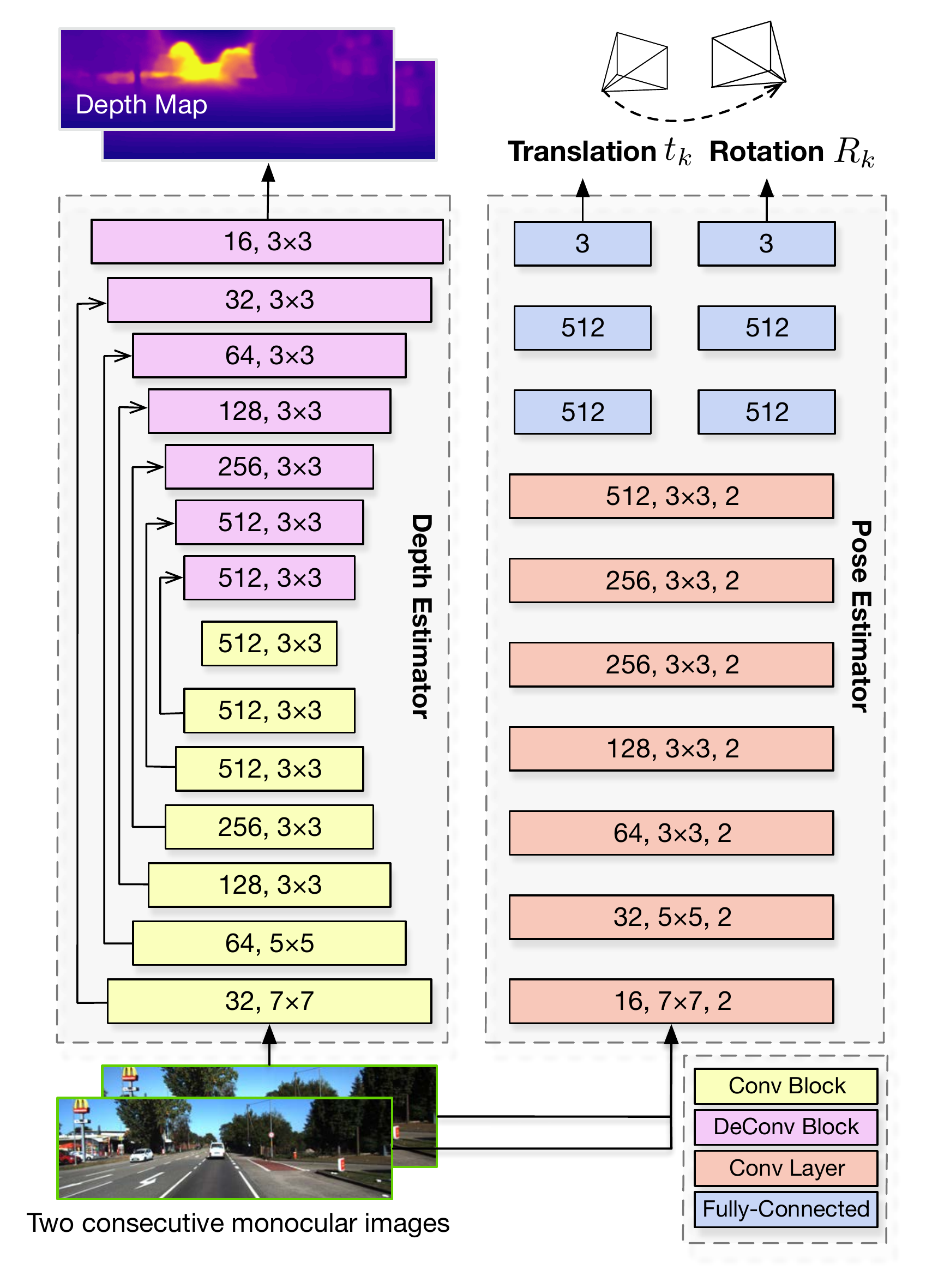}
  \caption{Architecture of our UnDeepVO.}
  \label{test_system_overview}
\end{figure}

Our system is composed of a pose estimator and a depth estimator, as shown in Fig. \ref{test_system_overview}. Both estimators take consecutive monocular images as inputs, and produce scaled 6-DoF pose and depth as outputs, respectively.

For the pose estimator, it is a VGG-based \cite{Simonyan15} CNN architecture. It takes two consecutive monocular images as input and predicts the 6-DoF transformation between them. Since rotation (represented by Euler angles) has high nonlinearity, it is usually difficult to train compared with translation. For supervised training, a popular solution is to give a bigger weight to the rotational loss \cite{kendall2015posenet} as a way of normalization. In order to better train the rotation with unsupervised learning, we decouple the translation and the rotation with two separate groups of fully-connected layers after the last convolutional layer. This enables us to introduce a weight normalizing the rotation and the translation predictions for better performance. The specific architecture of the pose estimator is shown in Fig. \ref{test_system_overview}. 

The depth estimator is mainly based on an encoder-decoder architecture to generate dense depth maps. Different from other depth estimation methods \cite{monodepth17,zhou2017unsupervised} which produce disparity images (inverse of the depth) from the network, the depth estimator of UnDeepVO is designed to directly predict depth maps. This is because training trails report that the whole system is easier to converge when training in this way.

For most monocular VO methods, a predefined scale has to be applied. One feature of our UnDeepVO is to recover absolute scale of 6-DoF pose and depth, it is credited to our training scheme shown in Fig. \ref{fig:TrainingScheme}. During training, we feed both left images and right images into the networks, and get 6-DoF poses and depths of left sequences and right sequences, respectively. We then use the input stereo images, estimated depth images and 6-DoF poses to construct the loss function considering the geometry of a stereo rig. 

As shown in Fig. \ref{fig:TrainingScheme}, we utilize both spatial and temporal geometric consistencies of a stereo image sequence to formulate the loss function. The red points in one image all have the corresponding ones in another. Spatial geometric consistency represents the geometric projective constraint between the corresponding points in left-right image pairs, while temporal geometric consistency represents the geometric projective constraint between the corresponding points in two consecutive monocular images (more details in section \uppercase\expandafter{\romannumeral4}). By using these constraints to construct the loss function and minimizing them all together, the UnDeepVO learns to estimate scaled 6-DoF poses and depth maps in an end-to-end unsupervised manner.

%%%%%%%%%%%%%%%%%%%%%%%%%%%%%%%%%%%%%%%%%%%%%%%%%%%%%%%%%%%%%%%%%%%%%
\section{Objective Losses for Unsupervised Training}

	\begin{figure}
		\centering
		\includegraphics[width=\columnwidth]{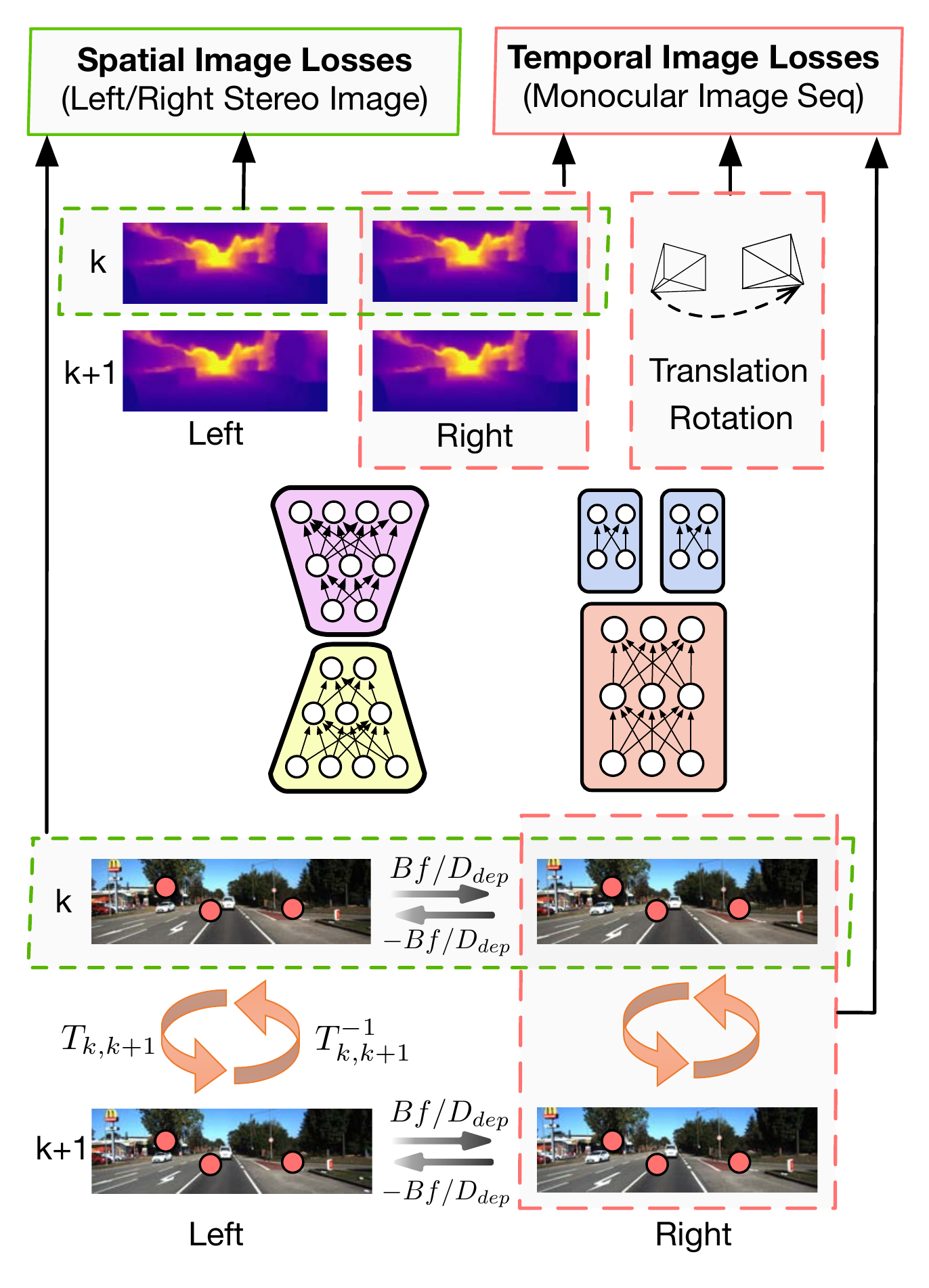}
		\caption{Training scheme of UnDeepVO. The pose and depth estimators take stereo images as inputs to estimate 6-DoF poses and depth maps, respectively. The total loss including spatial losses and temporal losses can then be calculated based on raw RGB images, estimated depth maps and poses.}
		\label{fig:TrainingScheme}
	\end{figure}

UnDeepVO is trained with losses through backpropagation. Since the losses are built on geometric constraints rather than labeled data, UnDeepVO is trained in an unsupervised manner. Its total loss includes spatial image losses and temporal image losses, as shown in Fig. \ref{fig:TrainingScheme}. The spatial image losses drive the network to recover scaled depth maps by using stereo image pairs, while the temporal image losses are designed to minimize the errors on camera motion by using two consecutive monocular images.

\subsection{Spatial Image Losses of a Stereo Image Pair}

The spatial losses are constructed from the geometric constraints between left and right stereo images. It is composed of left-right photometric consistency loss, disparity consistency loss and pose consistency loss. UnDeepVO relies on these losses to recover the absolute scale for the monocular VO.
	
\subsubsection{Photometric Consistency Loss}
The left-right projective photometric error is used as photometric consistency loss to train the network. Specifically, for the overlapped area between two stereo images, every pixel in one image can find its correspondence in the other with horizontal distance $D_{p}$ \cite{garg2016unsupervised}. Assume $p_{l}(u_{l},v_{l})$ is a pixel in the left image and $p_{r}(u_{r},v_{r})$ is its corresponding pixel in the right image. Then, we have the spatial constraint $u_{l}=u_{r}$ and $v_{l}=v_{r}+D_{p}$. The distance $D_{p}$ can be calculated by 
\begin{equation}
    D_{p} = Bf/D_{dep}
\end{equation}
where $B$ is the baseline of the stereo camera, $f$ is the focal length and $D_{dep}$ is the depth value of the corresponding pixel. We can construct a $D_{p}$ map with the depths estimated from the network to define the constraints between the left and right images. With this spatial constraint and the calculated $D_{p}$ map, we could synthesize one image from the other through ``spatial transformer'' \cite{jaderberg2015spatial}. The combination of an L1 norm and an SSIM term \cite{zhao2015l2} is used to construct the left-right photometric consistency loss:
\begin{align}
		L_{pho}^{l} =  \lambda_{s} L_{}^{SSIM}(I_{l}, {I}'_{l}) + (1-\lambda_{s}) L^{l_{1}}(I_{l}, {I}'_{l}) \\
		L_{pho}^{r} =  \lambda_{s} L_{}^{SSIM}(I_{r}, {I}'_{r}) + (1-\lambda_{s}) L^{l_{1}}(I_{r}, {I}'_{r} ) 
	\end{align}
where $I_{l}, I_{r}$ are the original left and right images respectively, ${I}'_{l}$ is the synthesized left image from the original right image $I_{r}$, and ${I}'_{r}$ is the synthesized right image from the original left image $I_{l}$, $L^{SSIM}$ is the operation defined in \cite{wang2004image} with a weight $\lambda_{s}$, and $L^{l_{1}}$ is the L1 norm operation.
%TODO: simply notation
	
\subsubsection{Disparity Consistency Loss}
Similarly, the left and right disparity maps (inverse of depth) are also constrained by $D_{p}$. The disparity map UnDeepVO used is
\begin{equation}
		D_{dis} = D_{p}\times I_{W}
\end{equation}
where $I_{W}$ is the width of original image size. Denote the left and right disparity maps as $D_{dis}^{l}$ and $D_{dis}^{r}$, respectively. We can use $D_{p}$ to synthesize $D_{dis}^{l'}, D_{dis}^{r'}$ from $D_{dis}^{r}, D_{dis}^{l}$. Then, the disparity consistency losses are
\begin{align}
		L_{dis}^{l} = L^{l_{1}}( D_{dis}^{l}, D_{dis}^{l'} )  \\
		L_{dis}^{r} = L^{l_{1}}( D_{dis}^{r}, D_{dis}^{r'} ) 
\end{align}

\subsubsection{Pose Consistency Loss}
We use both left and right image sequences to predict the 6-DoF transformation of the camera separately during training. Ideally, these two predicted transformations should be basically identical. Therefore, we can penalize the difference between them by
	\begin{align}
	L_{pos} = \lambda_{p} L^{l_{1}}(\mathbf{x}'_{l}, \mathbf{x}'_{r} ) + \lambda_{o} L^{l_{1}} ({\varphi}'_{l}, {\varphi}'_{r} )
	\end{align}
where $\lambda_{p}$ is the left-right position consistency weight, $\lambda_{o}$ is the left-right orientation consistency weight, and $[\mathbf{x}'_{l}, {\varphi}'_{l}]$ and $[\mathbf{x}'_{r}, {\varphi}'_{r}]$ are the predicted poses from the left and right image sequences, respectively.

\subsection{Temporal Image Losses of Consecutive Monocular Images}

Temporal loss is defined according to the geometric constraints between two consecutive monocular images. It is an essential part to recover the 6-DoF motion of camera. It comprises photometric consistency loss and 3D geometric registration loss.
	
\subsubsection{Photometric Consistency Loss}
The photometric loss is computed from two consecutive monocular images. Similar to DTAM \cite{newcombe2011dtam}, in order to estimate 6-DoF transformation, the projective photometric error is employed as the loss to minimize. Denote $I_{k}$, $I_{k+1}$ as the $k$th and $(k+1)$th image frame, respectively, and $p_{k}(u_{k}, v_{k})$ as one pixel in $I_{k}$, and $p_{k+1}(u_{k+1}, v_{k+1})$ as the corresponding pixel in $I_{k+1}$. Then, we can derive $p_{k+1}$ from $p_{k}$ through
\begin{align}
	p_{k+1} = KT_{k,k+1}D_{dep}K^{-1}p_{k}
\end{align}
where $K$ is the camera intrinsics matrix, $D_{dep}$ is the depth value of the pixel in the $k$th frame, $T_{k, k+1}$ is the camera coordinate transformation matrix from the $k$th frame to the $(k+1)$th frame. We can synthesize ${I}'_{k}$ and ${I}'_{k+1}$ from $I_{k+1}$ and $I_{k}$ by using estimated poses and ``spatial transformer'' \cite{jaderberg2015spatial}. Therefore, the photometric losses between the monocular image sequence are
\begin{align}
	L_{pho}^{k} = \lambda_{s} L_{}^{SSIM}(I_{k}, {I}'_{k} ) + (1-\lambda_{s}) L^{l_{1}}(I_{k}, {I}'_{k} ) \ \ \ \ \ \    \\
	L_{pho}^{k+1} = \lambda_{s} L_{}^{SSIM}(I_{k+1}, {I}'_{k+1} ) + (1-\lambda_{s}) L^{l_{1}}(I_{k+1}, {I}'_{k+1} ) 
	\end{align}
	
\subsubsection{3D Geometric Registration Loss}
3D geometric registration loss is to estimate the transformation by aligning two point clouds, similar to the Iterative Closest Point (ICP) technique. Assume $P_{k}(x,y,z)$ is a point in the $k$th camera coordination. It can then be transformed to the $(k+1)$th camera coordination as ${P}'_{k}(x,y,z)$ by using $T_{k, k+1}$. Similarly, points in the $(k+1)$th frame can be transformed to $k$th frame. Then, the 3D geometric registration losses between two monocular images are
\begin{align}
	L_{geo}^{k} = L^{l_{1}}( P_{k}, {P}'_{k} ) \ \ \    \\
	L_{geo}^{k+1} = L^{l_{1}}( P_{k+1}, {P}'_{k+1} )
\end{align}

% \begin{equation}
%    \begin{split}
%	    L = \lambda_{p}^{lr} (L_{pho}^{l} + L_{pho}^{r}) + \lambda_{d} (L_{dis}^{l} + L_{dis}^{r}) + \lambda_{pos} L_{pos}  \\
%        + \lambda_{p}^{k,k+1} (L_{pho}^{k} + L_{pho}^{k+1}) + \lambda_{g} (L_{geo}^{k} + L_{geo}^{k+1}) + \lambda_{smo} L_{smo}
%    \end{split}
% \end{equation}

% Where $\lambda_{p}^{lr}$, $\lambda_{d}$, $\lambda_{pos}$, $\lambda_{p}^{k,k+1}$, $\lambda_{g}$ are the corresponding weights for losses. $L_{smo}$ is the smoothness loss with a weight $\lambda_{smo}$ to enhance the performance of depth estimation. 
    
% losses used by others
In summary, the final loss function of our system combines the previous spatial and temporal losses together. The left-right photometric consistency loss has been used in \cite{garg2016unsupervised} and \cite{monodepth17} to estimate depth map. \cite{zhou2017unsupervised} introduced the photometric loss of a monocular image sequence for ego-motion and depth estimation. However, to the best of our knowledge, UnDeepVO is the first to recover both scaled camera poses and depth maps by benefiting all these losses together with the 3D geometric registration and pose consistency losses. 

 \begin{figure}[!tp]
    \centering
	\subfigure[02]{
	\begin{minipage}{0.225\textwidth}
	\includegraphics[width=1.0\textwidth]{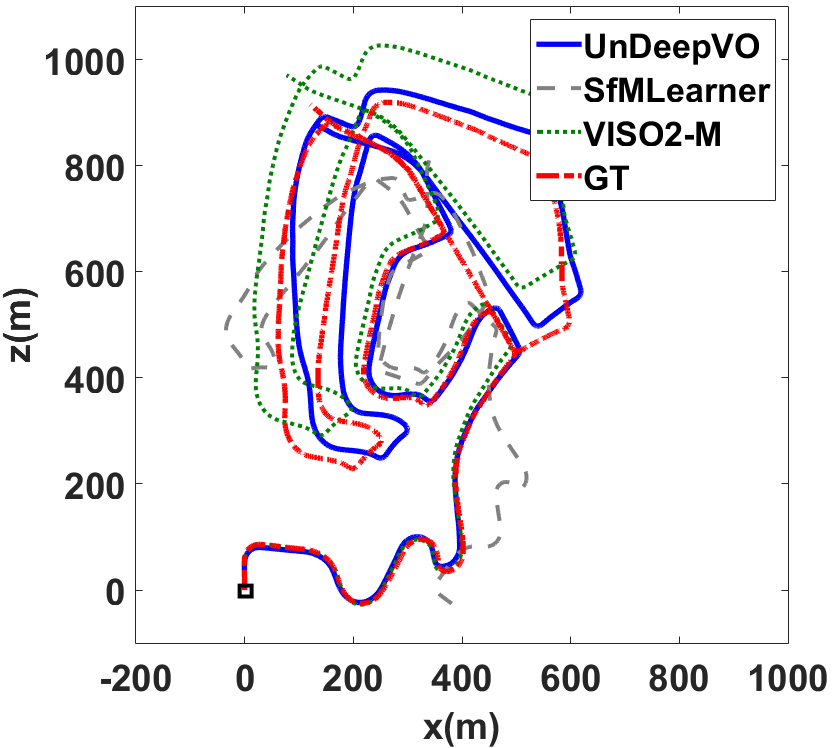} \\
	\end{minipage}
    }    
	\subfigure[05]{
	\begin{minipage}{0.225\textwidth}
	\includegraphics[width=1.0\textwidth]{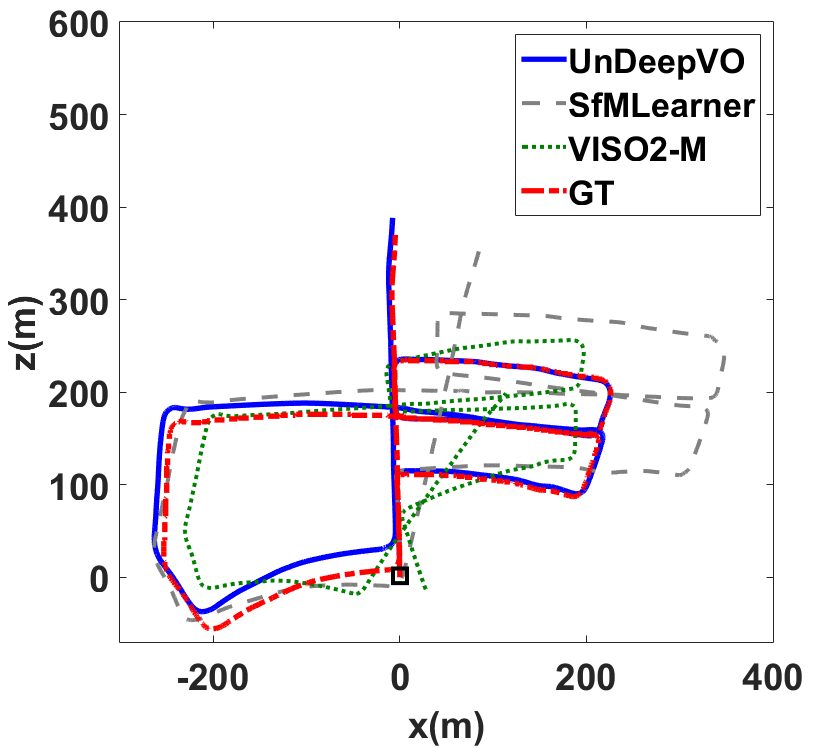} \\
	\end{minipage}
    }
	\subfigure[07]{
	\begin{minipage}{0.225\textwidth}
	\includegraphics[width=1.0\textwidth]{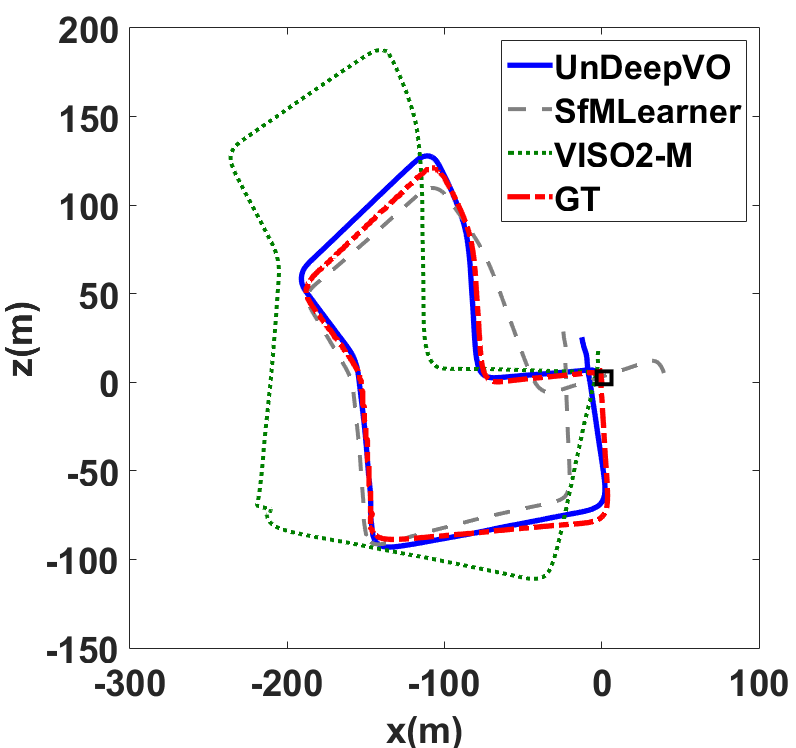} \\
	\end{minipage}
    }
	\subfigure[08]{
	\begin{minipage}{0.225\textwidth}
	\includegraphics[width=1.0\textwidth]{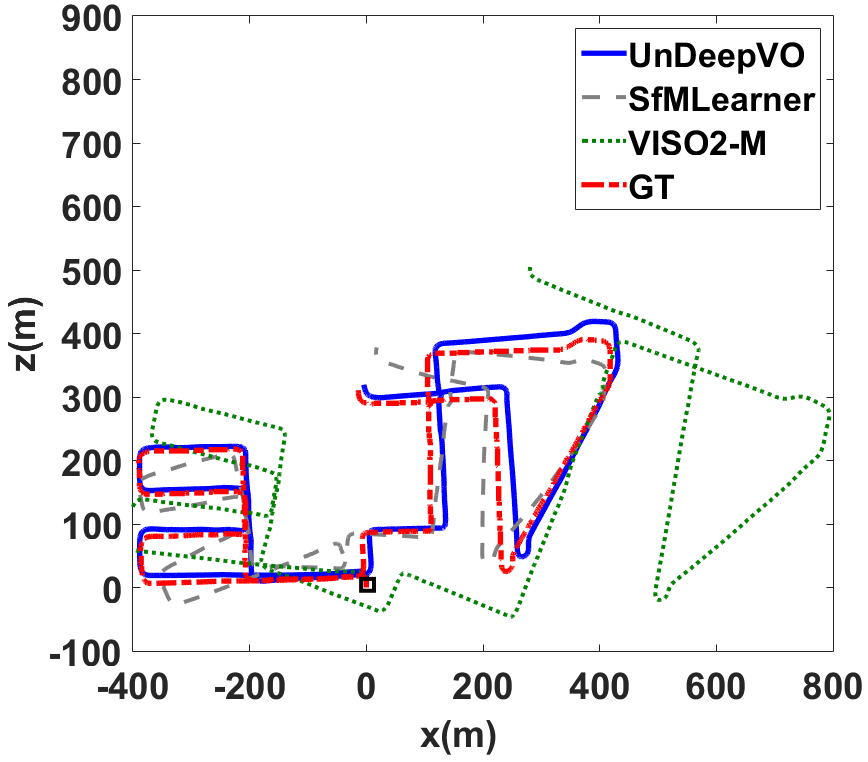} \\
	\end{minipage}
    }
    \caption{Trajectories of Sequence 02, 05, 07 and 08. Results of SfMLearner \cite{zhou2017unsupervised} are post-processed with 7-DoF alignment to ground truth since it cannot recover the scale. UnDeepVO and SfMLearner use images with size 416$\times$128. Images used by VISO2-M are 1242$\times$376.}
	\label{CNN_VO_00-08_fig}
\end{figure}

%%%%%%%%%%%%%%%%%%%%%%%%%%%%%%%%%%%%%%%%%%%%%%%%%%%%%%%%%%%%%%%%%%%%%%%%%%%%%%%%

\section{Experimental Evaluation}

%\subsection{Implementation Details}

In this section, we evaluated the proposed UnDeepVO system.\footnote{Video:\url{https://www.youtube.com/watch?v=5RdjO93wJqo&t}} The network models were implemented with the TensorFlow framework and trained with NVIDIA Tesla P100 GPUs. For testing, we used a laptop equipped with NVIDIA GeForce GTX 980M and Intel Core i7 2.7GHz CPU. The GPU memory needed for pose estimation is less than 400MB with 40Hz real-time performance.

	\begin{table*}% [bp]% [thpb] % [h]
		\centering
		\renewcommand{\arraystretch}{1.25}
		\caption{\small  VO results with our proposed UnDeepVO. All the methods listed in the table did not use loop closure. Note that monocular VISO2-M and ORB-SLAM-M (without loop closure) did not work with image size 416 $\times$ 128, the results were obtained with image size 1242$\times$376. 7-DoF (6-DoF + scale) alignment with the ground-truth is applied for SfMLearner and ORB-SLAM-M.}
		\label{CNN_VO_00-08_table}
		\small
		\begin{tabular*}{0.8\textwidth}{c@{\extracolsep{\fill}}cc||cc|cc|cc}
			\cline{1-9}
			\multirow{3}{*}{Seq.}   & \multicolumn{2}{c}{UnDeepVO}    & \multicolumn{2}{c}{SfMLearner \cite{zhou2017unsupervised}}                   & \multicolumn{2}{c}{VISO2-M}                  & \multicolumn{2}{c}{ORB-SLAM-M}        \\
            
			& \multicolumn{2}{c}{(416$\times$128)}    & \multicolumn{2}{c}{(416$\times$128)}                   & \multicolumn{2}{c}{(1242$\times$376)}                  & \multicolumn{2}{c}{(1242$\times$376)}        \\  \cline{2-9}
            
			& $t_{\text{rel}} {(\%)}$ & $r_{\text{rel}} (^{\circ})$          & $t_{\text{rel}} (\%)$ & $r_{\text{rel}} (^{\circ})$           & $t_{\text{rel}} (\%)$ & $r_{\text{rel}} (^{\circ})$ 		& $t_{\text{rel}} (\%)$ & $r_{\text{rel}} (^{\circ})$   \\ \cline{1-9}
            
			\multicolumn{1}{c|}{00}   & \textbf{4.14}             & \multicolumn{1}{c||}{\textbf{1.92}} & 65.27             & \multicolumn{1}{c|}{6.23}  & 18.24             & \multicolumn{1}{c|}{2.69}  & 25.29    & \multicolumn{1}{c}{7.37}     \\
            
			\multicolumn{1}{c|}{02}   & 5.58             & \multicolumn{1}{c||}{2.44} & 57.59             & \multicolumn{1}{c|}{4.09}  & \textbf{4.37}            & \multicolumn{1}{c|}{\textbf{1.18}}  & {\texttimes}    & \multicolumn{1}{c}{\texttimes}  \\
            
			\multicolumn{1}{c|}{05}   & \textbf{3.40}             & \multicolumn{1}{c||}{\textbf{1.50}} & 16.76             & \multicolumn{1}{c|}{4.06}  & 19.22             & \multicolumn{1}{c|}{3.54}  & 26.01    & \multicolumn{1}{c}{10.62}    \\   
            
            \multicolumn{1}{c|}{07}   & \textbf{3.15}             & \multicolumn{1}{c||}{\textbf{2.48}} & 17.52             & \multicolumn{1}{c|}{5.38}  & 23.61             & \multicolumn{1}{c|}{4.11}  & 24.53    & \multicolumn{1}{c}{10.83}    \\
            
            \multicolumn{1}{c|}{08}   & \textbf{4.08}             & \multicolumn{1}{c||}{\textbf{1.79}} & 24.02             & \multicolumn{1}{c|}{3.05}  & 24.18            & \multicolumn{1}{c|}{2.47}  & 32.40    & \multicolumn{1}{c}{12.13}   \\  \cline{1-9}
            
			\multicolumn{1}{l|}{mean}   & \textbf{4.07}             & \multicolumn{1}{c||}{\textbf{2.02}} & 36.23             & \multicolumn{1}{c|}{4.56}  & 17.93             & \multicolumn{1}{c|}{2.80}  & 27.05    & \multicolumn{1}{c}{10.23}     \\            

		\end{tabular*}
		\begin{itemize}
			\scriptsize
			\item $t_{\text{rel}}$: average translational RMSE drift $(\%)$ on length of 100m-800m.
			\item $r_{\text{rel}}$: average rotational RMSE drift ($^{\circ}/100$m) on length of 100m-800m.
		\end{itemize}
	\end{table*}
    
Adam optimizer was employed to train the network for up to 20-30 epochs with parameter $\beta_{1}=0.9$ and $\beta_{2}=0.99$. The learning rate started from 0.001 and decreased by half for every 1/5 of total iterations. The sequence length of images feeding to the pose estimator was 2. The size of image input to the networks was $416\times128$. We also resized the output images to a higher resolution to compute the losses and fine-tuned the networks in the end. Different kinds of data augmentation methods were used to enhance the performance and mitigate possible overfitting, such as image color augmentation, rotational data augmentation and left-right pose estimation augmentation. Specifically, we randomly selected 20\% images for color augmentation with random brightness in range [0.9, 1.1], random gamma in range [0.9, 1.1] and random color shifts in range [0.9, 1.1]. For rotational data augmentation, we increased the proportion of rotational data to achieve better performance in rotation estimation. Pose estimation consistency of left-right images was also used for left-right pose estimation augmentation.

 \begin{figure}[!t]
	\centering
	\subfigure[13]{
	\begin{minipage}{0.45\columnwidth}
	\includegraphics[width=1\columnwidth]{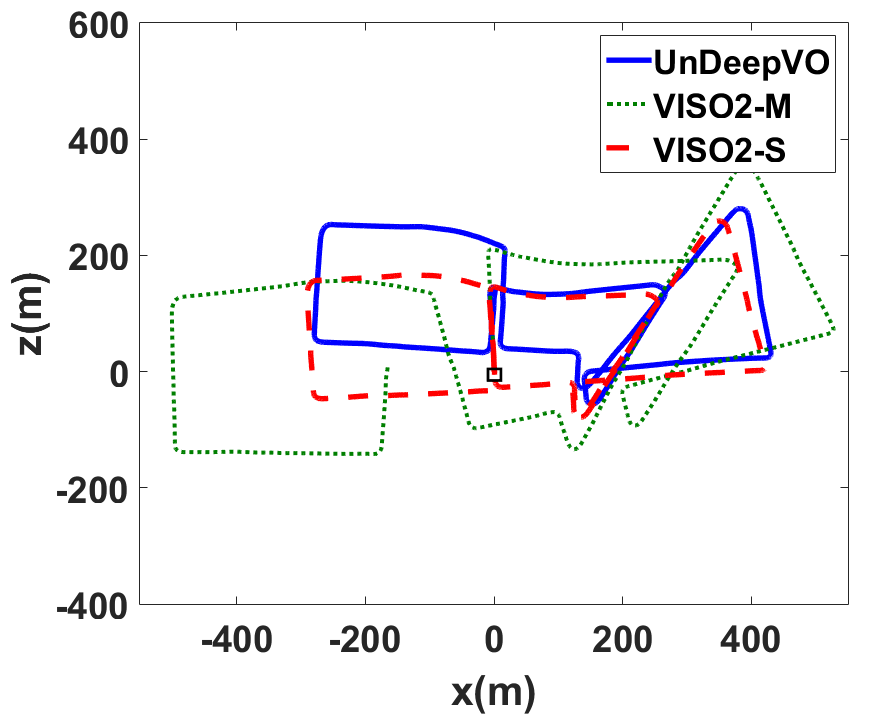} \\
	\end{minipage}
	}
	\subfigure[14]{
	\begin{minipage}{0.45\columnwidth}
	\includegraphics[width=1\columnwidth]{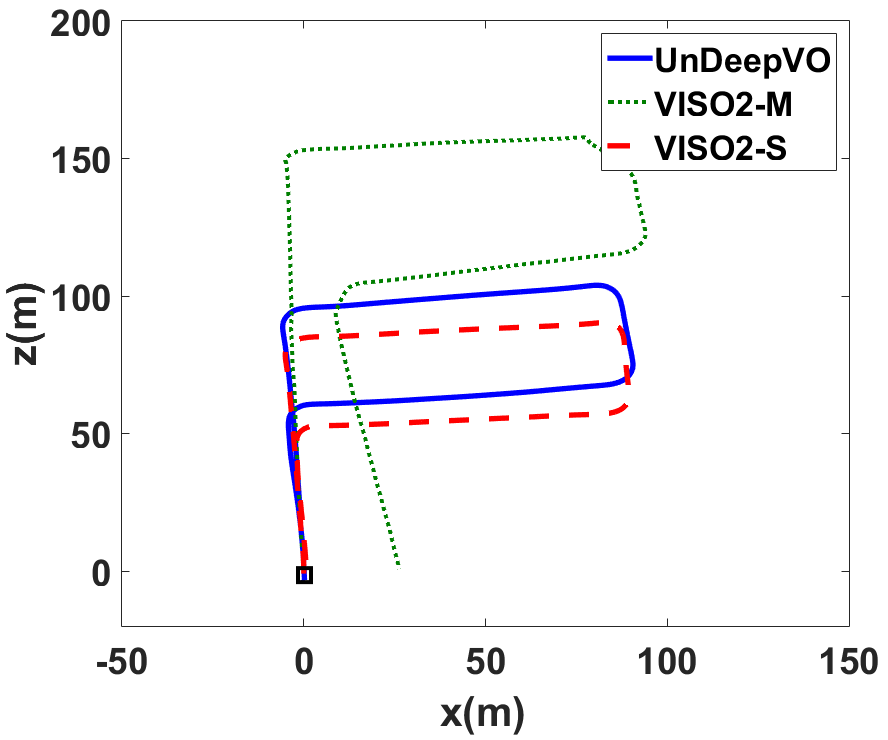} \\
	\end{minipage}
    }
	\subfigure[15]{
	\begin{minipage}{0.22\textwidth}
	\includegraphics[width=1\textwidth]{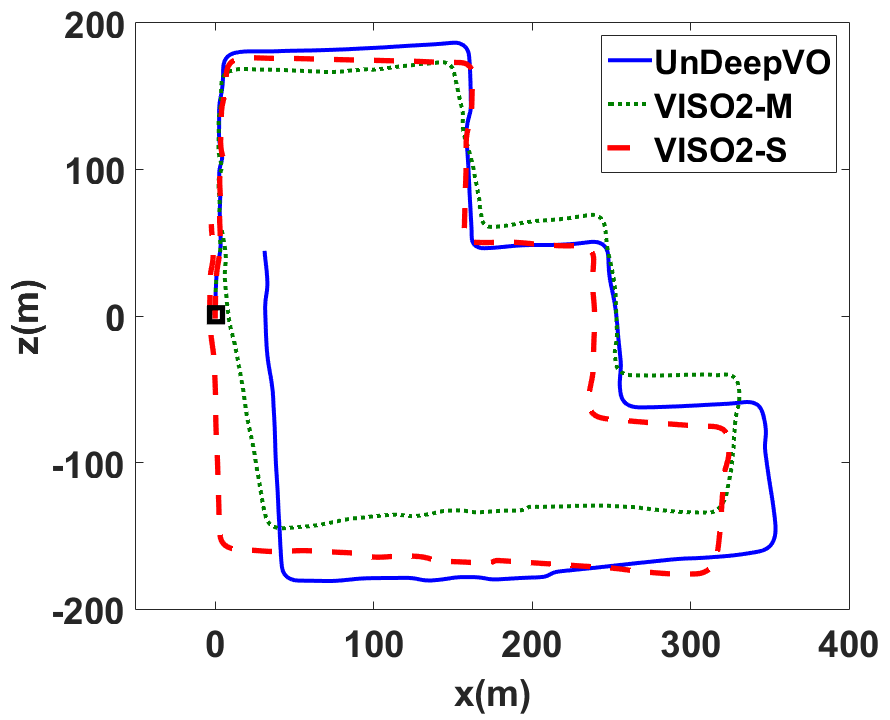} \\
	\end{minipage}
    }
	\subfigure[16]{
	\begin{minipage}{0.22\textwidth}
	\includegraphics[width=1\textwidth]{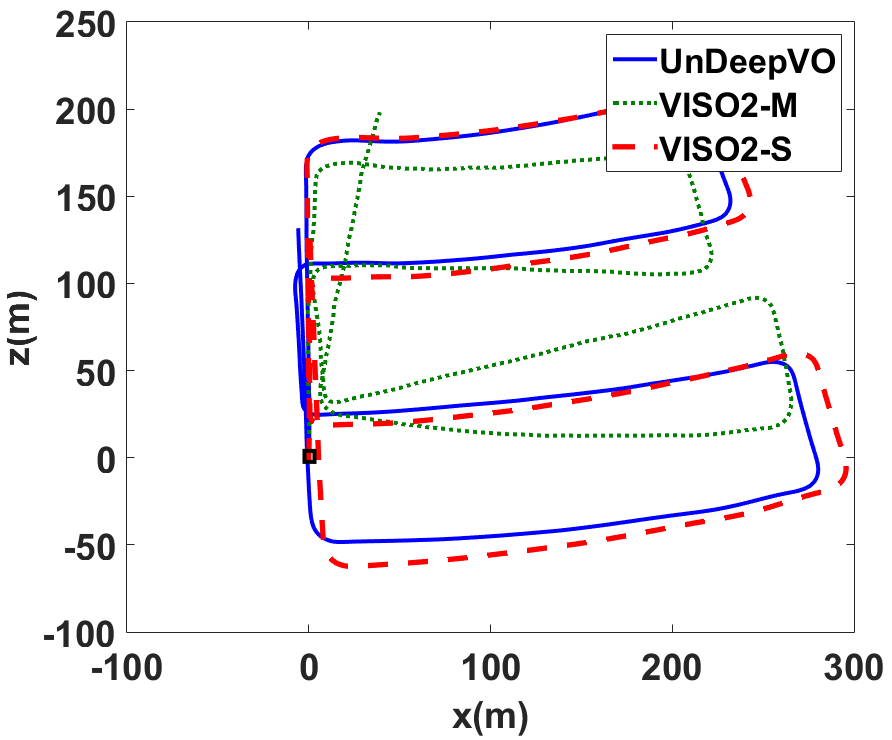} \\
	\end{minipage}
    }
	\subfigure[18]{
	\begin{minipage}{0.22\textwidth}
	\includegraphics[width=1\textwidth]{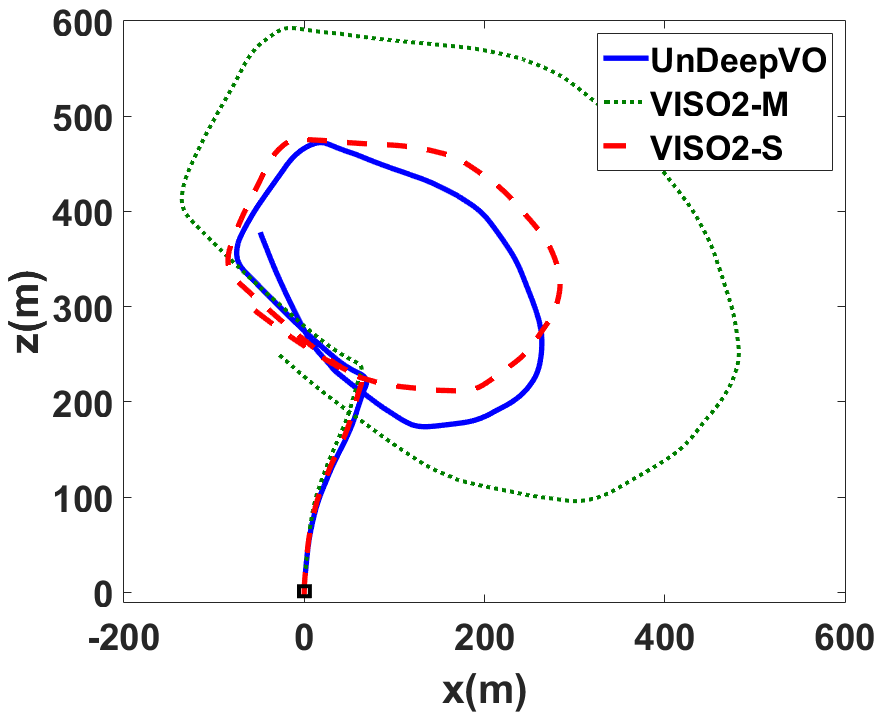} \\
	\end{minipage}
    }
	\subfigure[19]{
	\begin{minipage}{0.22\textwidth}
	\includegraphics[width=1\textwidth]{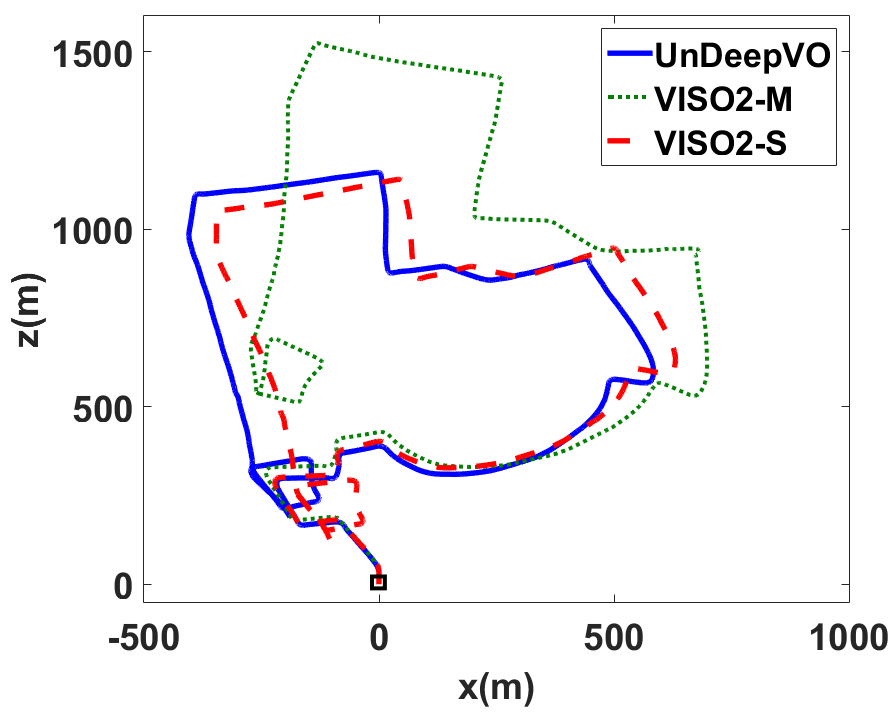} \\
	\end{minipage}
    }
    \caption{\small Trajectories of KITTI dataset with our UnDeepVO. No ground truth of poses is available for these sequences. Trajectories with both monocular VISO2-M and stereo VISO2-S are plotted. Our UnDeepVO works well on these sequences and is comparable to VISO2-S.}
	\label{CNN_VO_11_21_fig}
\end{figure}

\subsection{Trajectory Evaluation}
%TODO: add evaluation curves?

We adopted the popular KITTI Odometry Dataset\cite{Geiger2012CVPR} to evaluate the proposed UnDeepVO system, and compared the results with SfMLearner \cite{zhou2017unsupervised}, monocular VISO2-M and ORB-SLAM-M (without loop closure). In order to implement fair qualitative and quantitative comparison, we used the same training data as in SfMLearner \cite{zhou2017unsupervised} (sequences: 00-08). The trajectories produced by different methods are shown in Fig. \ref{CNN_VO_00-08_fig}, the comparison here shows the goodness of the network fit and is meaningful for structure-from-motion problem. Note that all the methods took monocular images for testing, and we post-process the scales for SfMLearner and ORB-SLAM-M as they cannot recover the scale of pose and depth. VISO2-M employed the fixed camera height for scale recovery. For ORB-SLAM-M, we disabled the local mapping and loop closure in order to perform VO only for comparison. The KITTI Odometry Dataset only provides the ground-truth of 6-DoF poses for Sequence 00-10. As shown in Fig. \ref{CNN_VO_00-08_fig}, the trajectories of UnDeepVO are qualitatively closest to the ground truth among all the methods. For sequences 11-21, there is no ground-truth available, and the trajectories of our method and VISO2-M are given in Fig. \ref{CNN_VO_11_21_fig}. The results of stereo VISO2-S (image resolution 1242 $\times$ 376) are provided for reference. As shown in the figure, our system's performance is comparable to that of VISO2-S.

%% merge ?
%\subsection{Pose Estimation Evaluation}
The detailed results (shown in Fig. \ref{CNN_VO_00-08_fig}) are listed in Table \ref{CNN_VO_00-08_table} for quantitative evaluation. We use the standard evaluation method provided along with KITTI dataset. Average translational root-mean-square error (RMSE) drift (\%) and average rotational RMSE drift ($^{\circ}/100$m) on length of 100m-800m are adopted. Since SfMLearner and ORB-SLAM-M cannot recover the scale of 6-DoF poses, we aligned their poses to the ground-truth with 6-DoF and scale (7-DoF). For monocular VISO2-M and ORB-SLAM without loop closure, they can not work with our input settings (image resolution $416\times128$), so we provide the results of both system with high resolution $1242 \times 376$. All the methods here did not use any loop closure technology. As shown in Table \ref{CNN_VO_00-08_table}, our method achieves good pose estimation performance among the monocular methods even with low resolution images and without the scale post-processing.  

\subsection{Depth Estimation Evaluation}
\begin{figure}
		\centering
		\framebox{\parbox{3.2in}{ \includegraphics[width=0.95\columnwidth]{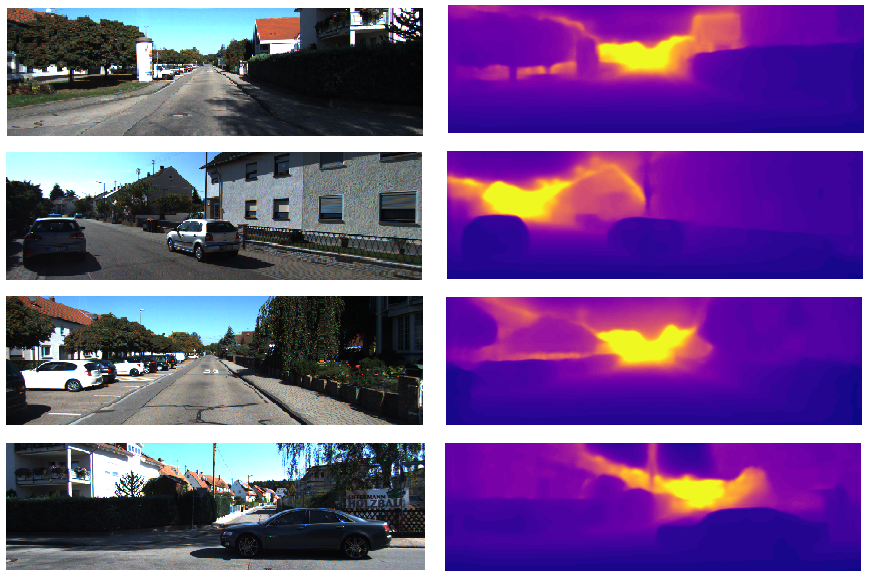}}}
		\caption{Depth images produced by our depth estimator. The left column are raw RGB images, and the right column are the corresponding depth images estimated.}
		\label{depth_map}
	\end{figure}

Our system can also produce the scaled depth map by using the depth estimator. Fig. \ref{depth_map} shows some raw RGB images and their corresponding depth images estimated from our system. As shown in Fig. \ref{depth_map}, the different depths of cars and trees are explicitly estimated, even the depth of trunks is predicted successfully. The detailed depth estimation results are listed in Table \ref{depth_comparison}. As shown in the table, our method outperforms the supervised one\cite{eigen2014depth} and the unsupervised one without scale \cite{zhou2017unsupervised}, but performs not as good as \cite{monodepth17}. This could be caused by a few reasons. First, we only used parts of KITTI dataset (KITTI odometry dataset) for training while all other methods use full KITTI dataset to train their networks. Second, \cite{monodepth17} used higher resolution ($512\times256$) input and a different net (ResNet-based architecture). Third, the temporal image sequence loss we used could introduce some noise (such as moving objects) for depth estimation.

\section{Conclusions}

In this paper, we presented UnDeepVO, a novel monocular VO system with unsupervised deep learning. The system makes use of spatial losses and temporal losses between stereo image sequences for unsupervised training. During testing, the proposed system can perform the pose estimation and dense depth map estimation with monocular images. Our system recovers the scale during the training stage, which distincts itself from other model based or learning based monocular VO methods. In general, unsupervised learning based VO methods have the potential to improve their performance with the increasing size of training datasets. In the next step, we will investigate how to train the UnDeepVO with large amount of datasets to improve its performance, such as robustness to image blurs, camera parameters, or illumination changes. In the future, we also plan to extend our system to a visual SLAM system to reduce the drift. Developing an unsupervised DeepVO system with stereo cameras or RGB-D cameras is also in consideration.

	\begin{table}%[bp]% [thpb] % [h]
		%\centering
		\renewcommand{\arraystretch}{1.25}
		\caption{Depth estimation results on KITTI using the split of Eigen et al.\cite{eigen2014depth}.}
		\label{depth_comparison}
		\footnotesize
		%\small
		\setlength{\tabcolsep}{3.5pt}
		% \scriptsize
		\begin{tabular*}{\columnwidth}{ccc||cccc}
			\cline{1-7}
			\multirow{2}{*}{Methods} & \multirow{2}{*}{Dataset}   & \multirow{2}{*}{Scale}  & \multicolumn{4}{c}{Error metric} \\ \cline{4-7}
			
			&	&      \multicolumn{1}{c||}{ }		& Abs Rel & Sq Rel  & RMSE & RMSE  log \\ \cline{1-7}
			
			\multicolumn{1}{c|}{Eigen \cite{eigen2014depth}}   & K (raw)         & {$\checked$}         & 0.214    & \multicolumn{1}{c}{1.605}  &   6.563      &    0.292 \\
			
			\multicolumn{1}{c|}{MonoDepth \cite{monodepth17}}   & K (raw)          & {$\checked$}        & 0.148    & \multicolumn{1}{c}{1.344}  &  5.927    & 0.247     \\
			
			\multicolumn{1}{c|}{SfMLearner \cite{zhou2017unsupervised}}   & K (raw)        & {\texttimes}             & 0.208    & \multicolumn{1}{c}{1.768}  & 6.856      & 0.283     \\  \cline{1-7}
			
			\multicolumn{1}{c|}{UnDeepVO}   & K (odo)         & {$\checked$}          &    0.183   & \multicolumn{1}{c}{1.73 }  &   6.57     &   0.268    \\
			
		\end{tabular*}
	\end{table}

%%%%%%%%%%%%%%%%%%%%%%%%%%%%%%%%%%%%%%%%%%%%%%%%%%%%%%%%%%%%%%%%%%%%%%%%%%%%%%%%


%%%%%%%%%%%%%%%%%%%%%%%%%%%%%%%%%%%%%%%%%%%%%%%%%%%%%%%%%%%%%%%%%%%%%%%%%%%%%%%%

\section*{ACKNOWLEDGMENT}

The authors would like to thank Robin Dowling for his support in experiments. The first author has been financially supported by scholarship from China Scholarship Council.

%%%%%%%%%%%%%%%%%%%%%%%%%%%%%%%%%%%%%%%%%%%%%%%%%%%%%%%%%%%%%%%%%%%%%

\addcontentsline{toc}{section}{References}
\bibliographystyle{IEEEtran}
\bibliography{IEEEabrv,references}

\begin{thebibliography}{10}
\providecommand{\url}[1]{#1}
\csname url@rmstyle\endcsname
\providecommand{\newblock}{\relax}
\providecommand{\bibinfo}[2]{#2}
\providecommand\BIBentrySTDinterwordspacing{\spaceskip=0pt\relax}
\providecommand\BIBentryALTinterwordstretchfactor{4}
\providecommand\BIBentryALTinterwordspacing{\spaceskip=\fontdimen2\font plus
\BIBentryALTinterwordstretchfactor\fontdimen3\font minus
  \fontdimen4\font\relax}
\providecommand\BIBforeignlanguage[2]{{%
\expandafter\ifx\csname l@#1\endcsname\relax
\typeout{** WARNING: IEEEtran.bst: No hyphenation pattern has been}%
\typeout{** loaded for the language `#1'. Using the pattern for}%
\typeout{** the default language instead.}%
\else
\language=\csname l@#1\endcsname
\fi
#2}}

\bibitem{davison2007monoslam}
A.~J. Davison, I.~D. Reid, N.~D. Molton, and O.~Stasse, ``{MonoSLAM}:
  {Real-time} single camera {SLAM},'' \emph{IEEE Transactions on Pattern
  Analysis and Machine Intelligence}, vol.~29, no.~6, pp. 1052--1067, 2007.

\bibitem{klein2007parallel}
G.~Klein and D.~Murray, ``Parallel tracking and mapping for small {AR}
  workspaces,'' in \emph{Mixed and Augmented Reality, 2007. ISMAR 2007. 6th
  IEEE and ACM International Symposium on}.\hskip 1em plus 0.5em minus
  0.4em\relax IEEE, 2007, pp. 225--234.

\bibitem{mur2015orb}
R.~Mur-Artal, J.~Montiel, and J.~D. Tardos, ``{ORB}-{SLAM}: {a} versatile and
  accurate monocular {SLAM} system,'' \emph{IEEE Transactions on Robotics},
  vol.~31, no.~5, pp. 1147--1163, 2015.

\bibitem{newcombe2011dtam}
R.~A. Newcombe, S.~J. Lovegrove, and A.~J. Davison, ``{DTAM}: {Dense} tracking
  and mapping in real-time,'' in \emph{Proceedings of the IEEE International
  Conference on Computer Vision (ICCV)}.\hskip 1em plus 0.5em minus 0.4em\relax
  IEEE, 2011, pp. 2320--2327.

\bibitem{engel2014lsd}
J.~Engel, T.~Sch{\"o}ps, and D.~Cremers, ``{LSD-SLAM}: {Large-scale} direct
  monocular {SLAM},'' in \emph{European Conference on Computer Vision
  (ECCV)}.\hskip 1em plus 0.5em minus 0.4em\relax Springer, 2014, pp. 834--849.

\bibitem{engel2017direct}
J.~Engel, V.~Koltun, and D.~Cremers, ``Direct sparse odometry,'' \emph{IEEE
  Transactions on Pattern Analysis and Machine Intelligence}, 2017.

\bibitem{kendall2015posenet}
A.~Kendall, M.~Grimes, and R.~Cipolla, ``{PoseNet}: {A} convolutional network
  for real-time 6-{DOF} camera relocalization,'' in \emph{Proceedings of the
  IEEE International Conference on Computer Vision (ICCV)}, 2015, pp.
  2938--2946.

\bibitem{li2017indoor}
R.~Li, Q.~Liu, J.~Gui, D.~Gu, and H.~Hu, ``Indoor relocalization in challenging
  environments with dual-stream convolutional neural networks,'' \emph{IEEE
  Transactions on Automation Science and Engineering}, 2017.

\bibitem{clark2017vidloc}
R.~Clark, S.~Wang, A.~Markham, N.~Trigoni, and H.~Wen, ``{VidLoc}: {6-DoF}
  video-clip relocalization,'' in \emph{Conference on Computer Vision and
  Pattern Recognition (CVPR)}, 2017.

\bibitem{costante2016exploring}
G.~Costante, M.~Mancini, P.~Valigi, and T.~A. Ciarfuglia, ``Exploring
  representation learning with {CNNs} for frame-to-frame ego-motion
  estimation,'' \emph{IEEE robotics and automation letters}, vol.~1, no.~1, pp.
  18--25, 2016.

\bibitem{wang2017deepvo}
S.~Wang, R.~Clark, H.~Wen, and N.~Trigoni, ``{DeepVO}: {Towards} end-to-end
  visual odometry with deep recurrent convolutional neural networks,'' in
  \emph{Robotics and Automation (ICRA), 2017 IEEE International Conference
  on}.\hskip 1em plus 0.5em minus 0.4em\relax IEEE, 2017, pp. 2043--2050.

\bibitem{ummenhofer2016demon}
B.~Ummenhofer, H.~Zhou, J.~Uhrig, N.~Mayer, E.~Ilg, A.~Dosovitskiy, and
  T.~Brox, ``{DeMoN}: {Depth} and motion network for learning monocular
  stereo,'' in \emph{Conference on Computer Vision and Pattern Recognition
  (CVPR)}, 2017.

\bibitem{clark2017vinet}
R.~Clark, S.~Wang, H.~Wen, A.~Markham, and N.~Trigoni, ``{VINet}:
  {Visual-Inertial} odometry as a sequence-to-sequence learning problem.'' in
  \emph{AAAI}, 2017, pp. 3995--4001.

\bibitem{pillai2017towards}
S.~Pillai and J.~J. Leonard, ``Towards visual ego-motion learning in robots,''
  \emph{arXiv preprint arXiv:1705.10279}, 2017.

\bibitem{jaderberg2015spatial}
M.~Jaderberg, K.~Simonyan, A.~Zisserman, \emph{et~al.}, ``Spatial transformer
  networks,'' in \emph{Advances in Neural Information Processing Systems},
  2015, pp. 2017--2025.

\bibitem{garg2016unsupervised}
R.~Garg, G.~Carneiro, and I.~Reid, ``{Unsupervised CNN} for single view depth
  estimation: {Geometry} to the rescue,'' in \emph{European Conference on
  Computer Vision (ECCV)}.\hskip 1em plus 0.5em minus 0.4em\relax Springer,
  2016, pp. 740--756.

\bibitem{monodepth17}
C.~Godard, O.~{Mac Aodha}, and G.~J. Brostow, ``Unsupervised monocular depth
  estimation with left-right consistency,'' in \emph{Conference on Computer
  Vision and Pattern Recognition (CVPR)}, 2017.

\bibitem{zhou2017unsupervised}
T.~Zhou, M.~Brown, N.~Snavely, and D.~G. Lowe, ``Unsupervised learning of depth
  and ego-motion from video,'' in \emph{Conference on Computer Vision and
  Pattern Recognition (CVPR)}, 2017.

\bibitem{Simonyan15}
K.~Simonyan and A.~Zisserman, ``Very deep convolutional networks for
  large-scale image recognition,'' in \emph{International Conference on
  Learning Representations (ICLR)}, 2015.

\bibitem{zhao2015l2}
H.~Zhao, O.~Gallo, I.~Frosio, and J.~Kautz, ``Is {L2} a good loss function for
  neural networks for image processing?'' \emph{ArXiv e-prints}, vol. 1511,
  2015.

\bibitem{wang2004image}
Z.~Wang, A.~C. Bovik, H.~R. Sheikh, and E.~P. Simoncelli, ``Image quality
  assessment: {From} error visibility to structural similarity,'' \emph{IEEE
  Transactions on Image Processing}, vol.~13, no.~4, pp. 600--612, 2004.

\bibitem{Geiger2012CVPR}
A.~Geiger, P.~Lenz, and R.~Urtasun, ``Are we ready for autonomous driving? {The
  KITTI} vision benchmark suite,'' in \emph{Conference on Computer Vision and
  Pattern Recognition (CVPR)}, 2012.

\bibitem{eigen2014depth}
D.~Eigen, C.~Puhrsch, and R.~Fergus, ``Depth map prediction from a single image
  using a multi-scale deep network,'' in \emph{Advances in neural information
  processing systems}, 2014, pp. 2366--2374.

\end{thebibliography}

\end{document}